\pdfoutput=1

\documentclass[11pt]{article}

\usepackage[]{ACL2023}

\usepackage{times}
\usepackage{latexsym}
\usepackage{CJKutf8}
\usepackage[tone]{tipa}
\usepackage{tipa}
\usepackage{amsmath}
\usepackage{amssymb}
\usepackage{tikz}
\usepackage{ctable}
\usepackage{graphicx}
\usepackage{caption}
\usepackage{balance}
\usepackage{subfig}
\usepackage{hyperref}
\usepackage{arydshln}
\usepackage{balance}

\usepackage[T1]{fontenc}

\usepackage[utf8]{inputenc}
\usepackage[vietnamese,german,english]{babel}

\usepackage{microtype}

\usepackage{inconsolata}

\title{Enhancing Cross-lingual Transfer via Phonemic Transcription Integration}

\author{Hoang H. Nguyen$^{1}$, Chenwei Zhang$^{2}$, Tao Zhang$^{1}$, Eugene Rohrbaugh$^{3}$, Philip S. Yu$^{1}$\\
  $^1$ Department of Computer Science, University of Illinois at Chicago, Chicago, IL, USA \\
  $^2$ Amazon, Seattle, WA, USA \\
  $^3$ Harrisburg University of Science and Technology , Harrisburg, PA, USA \\
  \texttt{\{hnguy7,tzhang90,psyu\}@uic.edu, cwzhang@amazon.com, gene.rohrbaugh@gmail.com}
  }

\begin{document}
\maketitle
\begin{abstract}
Previous cross-lingual transfer methods are restricted to orthographic representation learning via textual scripts. This limitation hampers cross-lingual transfer and is biased towards languages sharing similar well-known scripts. To alleviate the gap between languages from different writing scripts, we propose \textbf{PhoneXL}, a framework incorporating phonemic transcriptions as an additional linguistic modality beyond the traditional orthographic transcriptions for cross-lingual transfer. Particularly, we propose unsupervised alignment objectives to capture (1) local one-to-one alignment between the two different modalities, (2) alignment via multi-modality contexts to leverage information from additional modalities, and (3) alignment via multilingual contexts where additional bilingual dictionaries are incorporated. We also release the first phonemic-orthographic alignment dataset on two token-level tasks (Named Entity Recognition  and Part-of-Speech Tagging) among the understudied but interconnected Chinese-Japanese-Korean-Vietnamese (CJKV) languages. Our pilot study reveals phonemic transcription provides essential information beyond the orthography to enhance cross-lingual transfer and bridge the gap among CJKV languages, leading to consistent improvements on cross-lingual token-level tasks over orthographic-based multilingual PLMs.\footnote{Our code and datasets are publicly available at \href{https://github.com/nhhoang96/phonemic\_xlingual}{https://github.com/nhhoang96/phonemic\_xlingual}}

\end{abstract}

\section{Introduction}
Despite recent advances in cross-lingual pre-trained language models (PLM) such as mBERT \cite{devlin2019bert}, XLM-R \cite{conneau-etal-2020-unsupervised}, PLMs remain heavily biased towards high-resourced languages due to the skewed amount of available pre-training data under parameter capacity constraints. This heavily affects the downstream task performance of less-represented languages during pre-training. In addition, as most high-resourced languages share the similar scripts (i.e. mostly Latin scripts), PLMs tend to perform well on languages sharing similar scripts with those languages \cite{pires2019multilingual, muller2021being, fujinuma-etal-2022-match}. As most challenging low-resourced languages do not share similar scripts with common high-resourced languages, this limitation leaves them significantly behind. 

\begin{table}[bt]
\centering
\caption{Orthographic and Phonemic representations\footnotemark \; of name entities across CJKV languages. \textcolor{blue}{Blue} and \textcolor{red}{Red} denote pre-segmented phrases that share similar meanings. 
}
\vspace*{-0.3cm}
\resizebox{\columnwidth}{!}{%
\begin{tabular}{|c||c|c|c|c||}
\hline 
  & \multicolumn{1}{c|}{Orthographic} & \multicolumn{1}{c|}{Phonemic} \\
 \hline
EN & \textcolor{blue}{electronic} \textcolor{red}{industry} 
& \textipa{\textcolor{blue}{IlEkt\*rAnIk} \textcolor{red}{Ind@st\*ri}} 
\\
 \hdashline
ZH (src) &  \begin{CJK*}{UTF8}{gbsn}
\textcolor{blue}{电子} \textcolor{red}{行业} 
\end{CJK*} & 
\textipa{\textcolor{blue}{tjEn tsW} \textcolor{red}{xAN jE}} 
\\

\fontencoding{T5}\selectfont
VI (tgt) & \fontencoding{T5}\selectfont{\textcolor{red}{Công nghiệp}} 

\fontencoding{T5}\selectfont{\textcolor{blue}{Điện tử}} & \textcolor{red}{\textipa{koN Ni@p}} \textcolor{blue}{\textipa{di@n tW}}
\\ 
\hdashline
JA (src) &
\begin{CJK*}{UTF8}{min}
\textcolor{blue}{電子} \textcolor{red}{産業}
\end{CJK*}
& \textipa{ \textcolor{blue}{dEnSi} \textcolor{red} {s\ae Nju}} 

\\
KO (tgt) & \begin{CJK*}{UTF8}{mj}
\textcolor{blue}{전자} \textcolor{red}{산업}
\end{CJK*} 
& \textipa{\textcolor{blue}{\t{\textdyoghlig}E@nj@} \textcolor{red}{s\ae niawp}}
\\

\hline
EN &  \textcolor{blue}{Vietnam} \textcolor{red}{News Agency} 
&  \textipa{\textcolor{blue}{viEtnAm} \textcolor{red}{nuz ej\t{\textdyoghlig}@nsi}} 
\\ \hdashline

ZH (src) & 
\begin{CJK*}{UTF8}{gbsn}
\textcolor{blue}{越南} \textcolor{red}{通讯社}
\end{CJK*} & 
\textipa{\textcolor{blue}{ 4\oe \;nan \textcolor{red}{t\super hUN  Cyn \:s7}}} 
\\ 

VI (tgt) & \fontencoding{T5}\selectfont{\textcolor{red}{Thông tấn xã }} 
\fontencoding{T5}\selectfont{\textcolor{blue}{Việt Nam}} & \textcolor{red}{\textipa{ t\super hoN t7n sa}} \textcolor{blue}{\textipa{vi@t }}
\textcolor{blue}{\textipa{nam}} 
\\ \hdashline
JA (src) &
\begin{CJK*}{UTF8}{min}
\textcolor{blue}{ベトナム} \textcolor{red}{通信社}
\end{CJK*}
& \textipa{ \textcolor{blue}{bIt@nAmu} \textcolor{red} {ts2uSInS@}}
\\

KO (tgt) & \begin{CJK*}{UTF8}{mj}
\textcolor{blue}{베트남} \textcolor{red}{통신사}
\end{CJK*} 
& \textipa{\textcolor{blue}{bEtun\ae m} \textcolor{red}{tONsIns@}} 
\\
\hline
\end{tabular}%
}
\label{tab:intro_ex}
\end{table}

\footnotetext{ For the sake of simplicity and clearer comparison of phonemic representation similarity between source and target languages, we omit the tonal IPA characters. Tonal IPA characters are preserved as a part of the phonemic inputs for tokenization and training purposes.}

To alleviate the challenges on low-resource or zero-resource target languages, recent works attempt to transfer knowledge from high-resourced languages (typically English (EN)) to low-resourced target languages via augmentations from bilingual dictionaries and parallel corpora. However, these approaches are restricted to English source language and results in less significant performance gain on languages that are more distant from English \cite{yang2022enhancing, fujinuma-etal-2022-match}. Languages can be considered distant due to the differences in orthography, phonology, morphology and grammatical structures. The fact that performance drop occurs on distant languages when transferring from English indicates that additional works are needed to exploit connectivity between closely related languages, especially under extreme parameter constraints. 

Besides purely relying on the orthographic representation in the format of written scripts, additional modality of languages such as articulatory signals can provide essential information beyond the written scripts to enhance textual representations \citep{bharadwaj-etal-2016-phonologically, chaudhary2018adapting}. Phonemic transcriptions which capture linguistic articulatory signals are beneficial to understanding non-Latin-based languages when integrated with PLMs \cite{sun-etal-2021-chinesebert}. They can also facilitate for knowledge transfer between languages sharing lexical similarities in phonology but possessing different writing scripts. As demonstrated in Table \ref{tab:intro_ex}, despite differences in orthographic representations, the terms ``\begin{CJK*}{UTF8}{gbsn}电子 \end{CJK*}'' (ZH) and ``{\fontencoding{T5}\selectfont Điện tử}'' (VI) possess significant phonemic similarities when encoded into International Phonetic Alphabet (IPA). Similarly, although ``\begin{CJK*}{UTF8}{min}ベトナム \end{CJK*}'' (JA) and ``\begin{CJK*}{UTF8}{mj}베트남 \end{CJK*}'' (KO) are different, their phonemic representations (``\textipa{bIt@nAmu}'' and ``\textipa{bEtun\ae m}'' respectively) are almost identical in terms of articulatory features. 

Motivated by the inherent lexical similarities in terms of phonology among CJKV languages, we propose a novel cross-lingual transfer framework to integrate and synthesize 
two specific linguistic modalities (1) textual orthographic input scripts, (2) phonemic transcription, represented in International Phonetic Alphabet (IPA) format. Our unified cross-lingual transfer framework aims to effectively
(1) align both orthographic and phonemic transcriptions via multi-modality learning, (2) capture additional alignment between the two modalities via contextual information, (3) enhance cross-lingual transfer of the two modalities with additional bilingual dictionary. Our work specifically targets Chinese-Vietnamese-Japanese-Korean languages which are not well-studied in cross-lingual transfer and possess lexical similarities with one another in terms of phonology. Our contributions can be summarized as follows:
\begin{itemize}
    \item We provide the first pre-processed orthographic-phonemic transcription alignment dataset for token-level tasks (i.e. Part-of-Speech Tagging (POS) and Named Entity Recognition (NER)) among CJKV languages (Chinese-Japanese-Korean-Vietnamese). 
    \item We propose a multi-modality learning paradigm with unsupervised alignment objectives to fuse the knowledge obtained from both modalities/ transcriptions to enhance cross-lingual transfer.
    \item Our proposed framework yields consistent improvements over the orthographic-based multilingual PLMs (mBERT and XLM-R) on both POS and NER tasks.  
\end{itemize}
\section{Related Work}

\paragraph{Cross-lingual Transfer} 
Recent works in Cross-lingual transfer focus on generating multilingual contextualized representation for different languages based on the Pre-trained Language Models (PLM) via bilingual dictionaries \cite{qin2021cosda} and/or machine translation approaches \cite{fang2021filter, yang2022enhancing}. \citet{qin2021cosda} proposes a comprehensive code-switching technique via random selection of languages, sentences, and tokens to enhance multilingual representations, leading to improved performance on target languages on different downstream tasks. On the other hand, other approaches leverage parallel corpora generated by Machine Translation to (1) distill knowledge from source languages to target languages \cite{fang2021filter} or augment source language data with target language knowledge during training \cite{yang2022enhancing, zheng2021consistency}.   However, current cross-lingual efforts concentrate on single source language (EN) to multiple target languages. Under parameter capacity constraints, cross-lingual transfer has been shown to be biased towards high-resourced languages which share similar scripts and possess larger corpora of unlabeled data during pre-training \cite{fujinuma-etal-2022-match}. Unlike previous works, we specifically target enhancing performance of low-resourced languages by exploiting inherent linguistic similarities between closely-connected languages \cite{nguyen2019cross,zampieri2020natural}. 

\paragraph{Multi-modality Learning}
Multi-modality learning \cite{radford2021learning, li2021align, li2022blip} was initially proposed for the task of Visual-Question Answering \cite{goyal2017making}. The objective is to find alignment between the given images and textual input (i.e. caption). The two aligned modalities are trained to maximize the agreement with ground truth textual-image alignment. Despite its simple objectives, the CLIP \cite{radford2021learning} pre-training mechanism is considered the state-of-the-art in multi-modality representation learning.  Motivated by multi-modality learning, we integrate multi-modality learning approaches in unifying two modalities of transcriptions (orthographic and phonemic) for better representation enrichment.  
\begin{figure*}[bt]
    \centering
   \includegraphics[trim={0.0cm 0.0cm 0.0cm 0.0cm},clip, width=\textwidth]{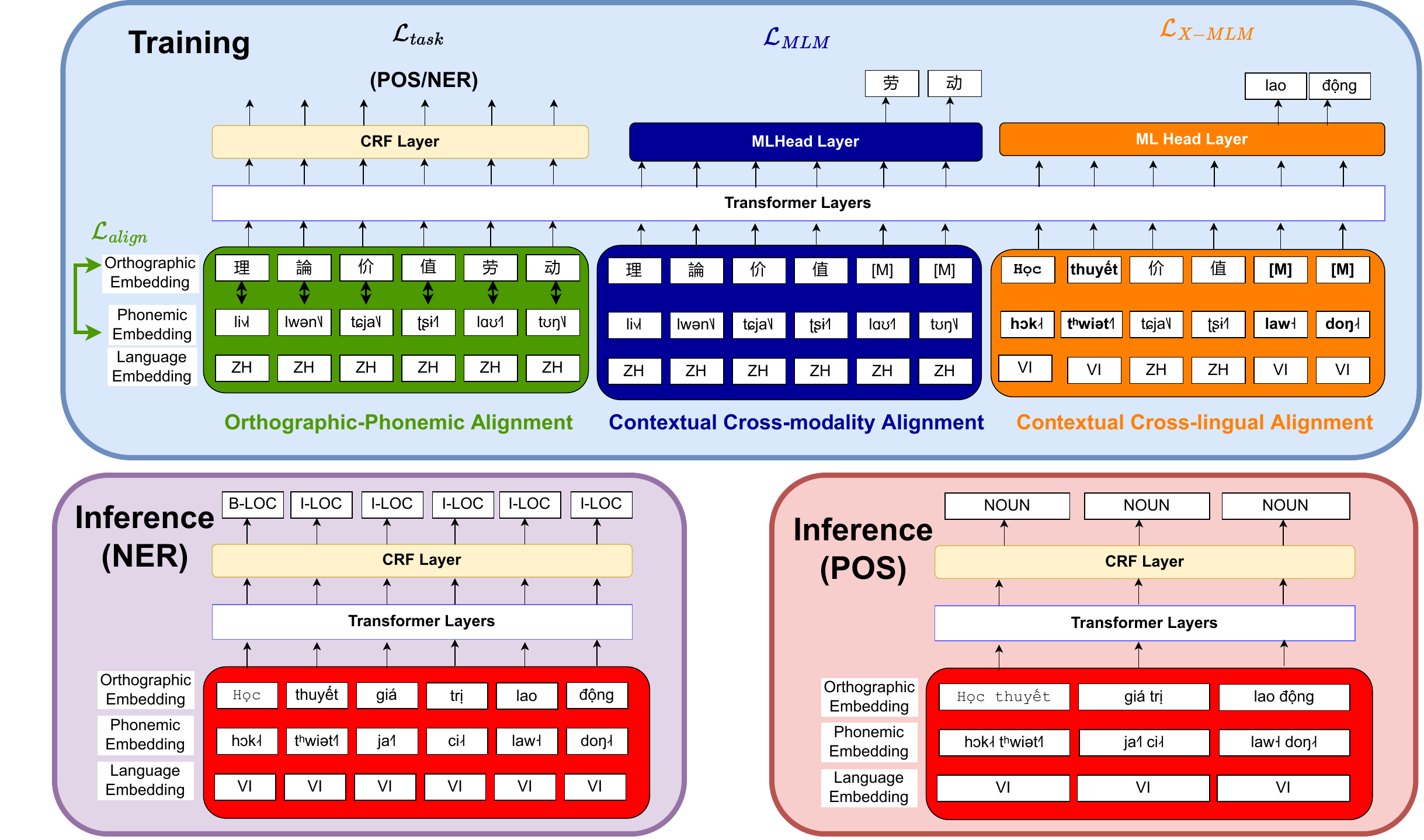}
    \caption{Illustration of the Proposed \textbf{PhoneXL} Model Overview. The model is trained with labeled source language data (ZH) and tested on target language data (VI). Orthographic embedding is identical to Token Embedding in PLMs and receives orthographic inputs. Phonemic Embedding maps phonemic inputs in IPA format to the embedding space.   \textcolor{green}{Green}, \textcolor{blue}{Blue}, \textcolor{orange}{Orange} denote \textcolor{green}{$\mathcal{L}_{align}$}, \textcolor{blue}{$\mathcal{L}_{MLM}$}, \textcolor{orange}{$\mathcal{L}_{XMLM}$} training objectives respectively. \textbf{Bold} represents target language words that are existent in bilingual dictionary to generate code-switching inputs. Parameters of Transformer Layers are shared across different inputs and unsupervised learning objectives.}
    \label{fig:overview}
    \vspace*{-0.5cm}
\end{figure*}

\section{Problem Formulation}
In this work, we study the problem of Cross-lingual Transfer in a bilingual setting where there exists annotated data collection of high-resource language S, namely $D_{S}^{train} = \{(X_{i}^{(S)}, Y_{i}^{(S)})\}_{i=1}^{N_{s}}$, and unlabeled data collection of low-resource target language T, denoted as $D_{T}^{test} = \{(X_{j}^{(T)})\}_{j=1}^{N_{t}}$. $N_{s}, N_{t}$ denote the sample size of source language training data and target language inference dataset respectively. 


Formally, given an i-th input source language utterance with the length of $M$ orthographic tokens $x_{i}^{(S)}=[x_{i,1}^{(S)},x_{i,2}^{(S)}..., x_{i,M}^{(S)}]$ and the corresponding phonemic transcriptions $z_{i}^{(S)}=[z_{i,1}^{(S)},z_{i,2}^{(S)}..., z_{i,M}^{(S)}]$ and token-level labels $y_{i}^{(S)}=[y_{i,1}^{(S)},y_{i,2}^{(S)}..., y_{i,M}^{(S)}]$, the overall training objective is summarized as:
\begin{equation}
\label{eq:eq_intro}
 \theta_{S} = \underset{\theta}{\mathrm{argmin}}\frac{1}{N_{s}} \sum_{i=1} ^{N_{s}} l (F(x^{(S)}_{i},z^{(S)}_{i}; \theta), y_{i}^{(S)}) 
\end{equation}

where $F(\cdot)$ denotes the transformation function that takes an input of both $x_{i}^{(S)},z_{i}^{(S)}$ to output probability prediction of label $y_{i}^{(S)}$ for individual tokens. $\theta$ denotes the parameters of the transformation framework and $l(\cdot)$ is the token-level cross-entropy loss. 

The overall trained framework is then evaluated in a zero-shot setting on target language $T$ as follows:
\begin{equation}
 \begin{aligned}
    p(y^{(T)}|x^{(T)},z^{(T)}) = \underset{k}{arg max} F(x^{(T)},z^{(T)}; \theta_{S})
      \end{aligned}
\end{equation}
where $k$ denotes the label space of token-level tasks.
In our cross-lingual transfer settings, no labeled target data or parallel data between source and target language is leveraged during training.

\section{Proposed Framework}
 \label{sec:framework}
In this section, we introduce our proposed Cross-lingual Transfer Framework, namely \textbf{PhoneXL}, with 3 major learning objectives: (1) Orthographic-Phonemic Alignment ($\mathcal{L}_{align}$), (2) Contextual Cross-modality Alignment ($\mathcal{L}_{MLM}$) , (3) Contextual Cross-lingual Alignment ($\mathcal{L}_{XMLM}$). The overall learning objective is summarized as follows:
\begin{equation}
    \mathcal{L} = \mathcal{L}_{task} + \alpha \mathcal{L}_{align} + \beta \mathcal{L}_{MLM} + \gamma \mathcal{L}_{XMLM}
\end{equation}

where $\mathcal{L}_{task}$ denotes the corresponding downstream token-level task and $\lambda,\beta,\gamma$ correspond to weights of the respective losses for balanced loss integration. For $\mathcal{L}_{task}$ is computed based on the training objective in Equation \ref{eq:eq_intro} as we leverage the generic CRF layer on top of the sequence output from PLM to generate the probability distribution of each token over the token-level class labels.   

\subsection{Orthographic-Phonemic Alignment}
\label{sec:subalign}
Traditional PLMs such as BERT \cite{devlin2019bert}, XLM-R \cite{conneau-etal-2020-unsupervised} encode the pre-tokenized input texts into 3 separate trainable embeddings: (1) token embedding ($\overrightarrow{w_{t}}$), (2) positional embedding($\overrightarrow{w_{p}}$), (3) segment embedding ($\overrightarrow{w_{s}}$) where $\overrightarrow{w_{t}}, \overrightarrow{w_{p}}, \overrightarrow{w_{s}} \in \mathbb{R}^{D}$ and D denotes the hidden dimensions of corresponding PLM. In our work, we name the token embedding as orthographic embedding (OE) to distinguish from (1) phonemic embedding, (2) unified embedding from both phonemic and orthographic inputs. The overall representation of individual tokens is computed as a summation of three types of embedding $\overrightarrow{w}=\overrightarrow{w_{t}} + \overrightarrow{w_{p}} + \overrightarrow{w_{s}}$.  

 With the goal of enhancing textual representations via both orthographic and phonemic transcriptions, we introduce the Phonemic Embedding (PE) to directly capture phonemic transcriptions. Phonemic embedding, namely $\overrightarrow{w_{PE}} \in \mathbb{R}^{D}$, encodes the representations of phonemic transcription inputs. Phonemic Embedding is integrated with orthographic embedding, positional and segment embedding to form the token representations \footnote{To ensure the alignment between length-variable phonemic and orthographic input resulted from tokenization, we meanpool the embedding of sub-tokens of individual inputs to construct representation for token-level tasks.}. 

 Motivated by previous works \cite{conneau2019cross, chaudhary2020dict}, we introduce additional Language Embedding ($\overrightarrow{w_{l}} \in \mathbb{R}^{D}$) to encode the input language types. These signals are beneficial to training objectives in cross-lingual settings with code-switched inputs introduced in Section \ref{subsec:xlingual}.
 
 The final word representation for the PLM Encoder is 
 $\overrightarrow{w} = \overrightarrow{w_{t}} +  \overrightarrow{w_{p}} + \overrightarrow{w_{s}} + \overrightarrow{w_{PE}} + \overrightarrow{w_{l}} $. We denote $\overrightarrow{v} = Q(\overrightarrow{w})$ as the word representation produced by PLM where $Q(\cdot)$ denotes the PLM Encoder function. 

To encourage the alignment between the orthographic textual input and its corresponding phonemic representation, we leverage cross-modality alignment and propose the computation of the  phonemic-orthographic alignment loss:
 \begin{equation}
      \mathcal{L}_{OtoP} = CrossEntropy(sim_{OtoP}, labels) 
 \end{equation}

The similarity matrices between phonemic and orthographic inputs ($sim_{OtoP}$) is computed as: 
\begin{equation}
    sim_{OtoP} = \sum_{m}^{M}\frac{\overrightarrow{w_{m,PE}}}{||\overrightarrow{w_{m,PE}}||} * \frac{\overrightarrow{w_{m,t}}}{||\overrightarrow{w_{m,t}}||} * \tau 
\end{equation}
where $\tau$ denotes the learnable soft temperature parameter and $||\cdot||$ is L2-normalization. $\overrightarrow{w_{m,PE}}, \overrightarrow{w_{im,t}}$ denote OE and PE of the m-th token in a sentence of length M.

 Similarly to text-image alignment in cross-modality learning \cite{radford2021learning}, the alignment is computed as a bi-directional interaction between orthographic and phonemic transcriptions. Therefore, the overall alignment loss is as follows: 
\begin{equation}
    \mathcal{L}_{align} = (\mathcal{L}_{OtoP} + \mathcal{L}_{PtoO}) /2
\end{equation}

\subsection{Contextual Cross-modality Alignment}
\label{subsec:multimodal}
The introduced alignment from \ref{sec:subalign} is restricted to 1-1 alignment between IPAs and tokens. However, contexts can significantly affect alignment between IPA and tokens. For instance, the orthography of \begin{CJK*}{UTF8}{gbsn}
行
\end{CJK*} usually corresponds to \textipa{[C\=iN]}. However, the orthography of \begin{CJK*}{UTF8}{gbsn}
行业
\end{CJK*} corresponds to \textipa{[x\'AN j\`E}]


To overcome the challenges, we propose introducing additional Masked Language Modeling to further align the two modalities. In other words, we randomly mask $\mu \%$ of input orthographic tokens and train the models to predict the masked tokens via (1) contextual/ non-masked orthographic tokens , (2) all of the phonemic transcriptions (including those of masked tokens). This objective encourages the alignment between phonemic and orthographic inputs via contextual information from both modalities of languages. Specifically, given a masked orthographic input and its corresponding phonemic representation, the model aims at predicting the masked tokens correctly. The loss is summarized as follows:
\begin{equation}
    \mathcal{L}_{MLM} = - \sum_{j \in C} log(P(y_{j}|\overrightarrow{v_{j}};\theta)
\end{equation}

where $y_{j},\overrightarrow{v_{j}}$ denote the j-th location ground truth MLM label and input representation produced by PLM as introduced in \ref{sec:subalign} respectively. C denotes the number of masked tokens in the input training sentence. 
\subsection{Cross-lingual Contextual Alignment}
\label{subsec:xlingual}
Up to this point, the model does not leverage any specific cross-lingual signals for knowledge transfer between source and target languages. Therefore, we further introduce the Cross-lingual Contextual Alignment objective via bilingual dictionary. Similarly to Contextual Multi-modal Alignment as introduced in Section \ref{subsec:multimodal}, we leverage MLM objectives to encourage the recognition of source language orthographic tokens given the phonemic inputs and multilingual orthographic contexts. The major difference between XMLM and MLM is that the input to XMLM is the code-switched input utterances which contain a mix of tokens from both source and target languages. Specifically, following \cite{qin2021cosda}, we conduct random code-switching of tokens of the source input utterances with ratio of $r\%$ where $r$ is a hyperparameter. The code-switched inputs follow the same procedure of MLM as introduced in Section \ref{subsec:multimodal}. The XMLM training objective is summarized as follows:
\begin{equation}
    \mathcal{L}_{XMLM} = - \sum_{j \in C'} log(P(y_{j}|\overrightarrow{\Tilde{v}_{j}};\theta)
\end{equation}
where $\overrightarrow{\Tilde{v}_{j}}$ is the PLM representation of j-th token for the code-switched input sentences of source language and $C'$ is the number of masked tokens based on percentage of code-switched source language inputs. Depending on the selected tokens for code-switching and its corresponding target language tokens, the absolute values of $C$ and $C'$ are not necessarily the same since the number of tokens in source samples and code-switched samples might not be identical.

\begin{table}[tb]
\centering
\caption{Details of processed PANX and UDPOS datasets. We report statistics of source language training set and target language testing set for ZH-VI language pair.}
\resizebox{\columnwidth}{!}{%
\begin{tabular}{|c|c|c|c|c|}
\hline 
& \multicolumn{2}{c|}{\textbf{PANX}} & \multicolumn{2}{c|}{\textbf{UDPOS}} \\
\hline 
& \textbf{Source} & \textbf{Target} & \textbf{Source} & \textbf{Target} \\
\hline 
\# Labels & 7 & 7 & 18 & 18 \\
\# Samples & 20000 & 10000 & 13320 & 1710 \\
Avg Token Length & 25.88  &  21.47 & 20.88 & 10.97 \\ 
Avg Tokenized Orthographic Length & 25.88  &  21.47 & 32.15 & 25.92 \\ 
Avg Tokenized Phonemic Length & 47.61  &  45.03 & 59.71 & 67.94 \\
\hline
\end{tabular}%
}
 \vspace*{-0.6cm}
\label{tab:detailsdataset}
\end{table}

\begin{table*}[htb]
\centering
\caption{NER and POS Experimental Results on PANX and UDPOS \textbf{test} datasets respectively.}
\vspace*{-0.1cm}
\resizebox{\textwidth}{!}{%
\begin{tabular}{|c||c|c||c|c||c|c||c|c||}
\hline 
 \textbf{Model} & \multicolumn{4}{c||}{\textbf{PANX}} & \multicolumn{4}{c||}{\textbf{UDPOS}}\\
\specialrule{.1em}{0.05em}{.05em}
 & \multicolumn{2}{c||}{ZH->VI} & \multicolumn{2}{c||}{JA->KO} 
  & \multicolumn{2}{c||}{ZH->VI} & \multicolumn{2}{c||}{JA->KO} \\
  \hline 
  & Source (ZH) & \textbf{Target (VI)} & Source (JA) & \textbf{Target (KO)} & Source (ZH) & \textbf{Target (VI)} & Source (JA) & \textbf{Target (KO)} \\
\specialrule{.1em}{0.05em}{.05em}

mBERT 
& 78.10 $\pm$ 0.25 & 49.94 $\pm$ 1.44 & 69.85 $\pm$ 0.17 & 26.64 $\pm$ 0.17
& 89.93 $\pm$ 0.02 & 48.62 $\pm$ 0.66 & 86.24 $\pm$ 0.13 & 43.63 $\pm$ 1.28
 \\
 
\textbf{PhoneXL (full)}
& \textbf{80.42 $\pm$ 0.07} & \textbf{52.28 $\pm$ 0.98} & \textbf{72.90 $\pm$ 0.37} & \textbf{29.25 $\pm$ 0.59} 

& \textbf{90.53 $\pm$ 0.04} & \textbf{50.71 $\pm$ 0.40} & \textbf{90.00 $\pm$ 0.15} & \textbf{46.75 $\pm$ 0.09} 

\\

PhoneXL (w $\mathcal{L}_{align}$) 
& 79.71 $\pm$ 0.21 & 51.09 $\pm$ 0.42 & 72.01 $\pm$ 0.11 & 28.23 $\pm$ 0.32
& 90.42 $\pm$ 0.03 & 50.29 $\pm$ 0.13 & 89.56 $\pm$ 0.07 & 45.96 $\pm$ 0.47
\\

PhoneXL (w $\mathcal{L}_{MLM}$) 
 & 79.70 $\pm$ 0.17 & 50.23 $\pm$ 1.63 & 72.62 $\pm$ 0.02 & 27.90 $\pm$ 0.11 
 & 90.44 $\pm$ 0.03 & 50.49 $\pm$ 0.67 & 89.53 $\pm$ 0.16 & 45.94 $\pm$ 0.45 
\\

PhoneXL (w $\mathcal{L}_{XMLM}$) 
& 79.69 $\pm$ 0.15  &  50.83 $\pm$ 0.63 & 72.57 $\pm$ 0.57 & 28.85 $\pm$ 0.71
 & 90.40 $\pm$ 0.05 & 50.20 $\pm$ 1.63 & 89.63 $\pm$ 0.16 & 45.25 $\pm$ 0.73 
\\

\hline
\hline
XLM-R
& 75.31 $\pm$ 0.46 & 35.68 $\pm$ 0.66 & 66.31 $\pm$ 0.06 & 14.80 $\pm$ 0.97
& 91.28 $\pm$ 0.04 & 50.40 $\pm$ 0.51 & 89.94 $\pm$ 0.17 & 46.16 $\pm$ 0.24
 \\

\textbf{PhoneXL (full) }
& \textbf{77.00 $\pm$ 0.24} & \textbf{38.88 $\pm$ 0.15} & \textbf{69.02 $\pm$ 0.24} & \textbf{16.39 $\pm$ 0.13}
& \textbf{91.43 $\pm$ 0.24} & \textbf{52.73 $\pm$ 0.86} & \textbf{90.06 $\pm$ 0.04} & \textbf{48.82 $\pm$ 0.43}
\\

PhoneXL (w $\mathcal{L}_{align}$) 
& 76.41 $\pm$ 0.09 & 37.04 $\pm$ 0.68 & 68.76 $\pm$ 0.25 & 15.34 $\pm$ 0.13
& 91.39 $\pm$ 0.02 & 52.46 $\pm$ 0.17 & 90.01 $\pm$ 0.12 & 47.96 $\pm$ 0.62
\\

PhoneXL (w $\mathcal{L}_{MLM}$) 
& 76.70 $\pm$ 0.07 &  37.29 $\pm$ 0.34 & 67.62 $\pm$ 0.13 & 15.16 $\pm$ 0.58
& 91.14 $\pm$ 0.02 & 51.88 $\pm$ 1.53 & 90.02 $\pm$ 0.05 & 47.83 $\pm$ 0.39
\\

PhoneXL (w $\mathcal{L}_{XMLM}$) 
& 76.52 $\pm$ 0.15 & 37.15 $\pm$ 0.30 & 68.68 $\pm$ 1.39 & 15.89 $\pm$ 0.79
& 91.04 $\pm$ 0.05 & 51.15 $\pm$ 1.40 & 89.90 $\pm$ 0.37 & 47.85 $\pm$ 0.56

\\
\hline
\end{tabular}%
}
\label{tab:test_full}
\end{table*}

\begin{table*}[htb]
\centering
\caption{NER and POS Baseline Results on PANX and UDPOS \textbf{test} datasets respectively. \textbf{Dict} denotes the assumptions of available bilingual dictionary and \textbf{MT} refers to the assumptions of available Machine Translations between source and target languages. Cross-lingual Transfer methods leverage either Dict or MT or both.  }
\vspace*{-0.1cm}
\resizebox{\textwidth}{!}{%
\begin{tabular}{|c||c|c||c|c||c|c||c|c||c|c||}
\hline 
 \textbf{Model} & \multicolumn{2}{c||}{\textbf{Assumption}} & \multicolumn{4}{c||}{\textbf{PANX}} & \multicolumn{4}{c||}{\textbf{UDPOS}}\\
\specialrule{.1em}{0.05em}{.05em}
 &  Dict & MT & \multicolumn{2}{c||}{ZH->VI} & \multicolumn{2}{c||}{JA->KO} 
  & \multicolumn{2}{c||}{ZH->VI} & \multicolumn{2}{c||}{JA->KO} \\
  \hline 
  & & &  Source (ZH) & \textbf{Target (VI)} & Source (JA) & \textbf{Target (KO)} & Source (ZH) & \textbf{Target (VI)} & Source (JA) & \textbf{Target (KO)} \\
\specialrule{.1em}{0.05em}{.05em}

mBERT 
& &
& 78.10 $\pm$ 0.25 & 49.94 $\pm$ 1.44 & 69.85 $\pm$ 0.17 & 26.64 $\pm$ 0.17
& 89.93 $\pm$ 0.02 & 48.62 $\pm$ 0.66 & 86.24 $\pm$ 0.13 & 43.63 $\pm$ 1.28
 \\

CoSDA-ML 
& $\checkmark$ & 
& 78.48 $\pm$ 0.34 & 47.82 $\pm$ 1.43 & 70.42 $\pm$ 0.50 & 25.76 $\pm$ 1.75
& 89.76 $\pm$ 0.19 & 49.84 $\pm$ 0.49 & 87.63 $\pm$ 0.14 & 41.19 $\pm$ 1.16
\\

X-MIXUP 
& & $\checkmark$
&  78.87 $\pm$ 0.17 & \textbf{52.98 $\pm$ 0.05} &  68.10 $\pm$ 0.69 & 26.41 $\pm$ 1.06
& 89.41 $\pm$ 0.10 & 50.05 $\pm$ 0.95 & 87.58 $\pm$ 0.17 & \textbf{48.47 $\pm$ 0.37}
\\
\hline

\textbf{PhoneXL (full)} & \checkmark & 
& \textbf{80.42 $\pm$ 0.07} & 52.28 $\pm$ 0.98 & \textbf{72.90 $\pm$ 0.37} & \textbf{29.25 $\pm$ 0.59} 

& \textbf{90.53 $\pm$ 0.04} & \textbf{50.71 $\pm$ 0.40} & \textbf{90.00 $\pm$ 0.15} & 46.75 $\pm$ 0.09 
\\

\hline
\hline
XLM-R
& &
& 75.31 $\pm$ 0.46 & 35.68 $\pm$ 0.66 & 66.31 $\pm$ 0.06 & 14.80 $\pm$ 0.97
& 91.28 $\pm$ 0.04 & 50.40 $\pm$ 0.51 & 89.94 $\pm$ 0.17 & 46.16 $\pm$ 0.24
\\

FILTER 
& & $\checkmark$
& 72.55 $\pm$ 0.11 & 40.17 $\pm$ 1.35 & 62.92 $\pm$ 0.26 & 18.60 $\pm$ 1.02 
& 90.57 $\pm$ 0.05 & \textbf{55.85 $\pm$ 0.27} & \textbf{90.81 $\pm$ 0.19} & 43.25 $\pm$ 1.52
\\

xTune 
& $\checkmark$ & $\checkmark$
& \textbf{77.48 $\pm$ 0.08} & \textbf{40.94 $\pm$ 0.87} & 68.02 $\pm$ 0.26 & \textbf{21.95 $\pm$ 1.02} 

& \textbf{91.75 $\pm$ 0.10} & 51.91 $\pm$ 0.74 & 89.75 $\pm$ 0.31 & \textbf{51.03 $\pm$ 1.26}

\\
X-MIXUP 
& & $\checkmark$
& 75.89 $\pm$ 0.46 & 38.22 $\pm$ 0.72 & 65.33 $\pm$ 0.69 & 16.43 $\pm$ 2.98
& 90.67 $\pm$ 0.06 & 50.30 $\pm$ 1.23 & 88.48 $\pm$ 0.23 & 50.63 $\pm$ 0.95
\\

\hline 
\textbf{PhoneXL (full)}
& $\checkmark$ &
& 77.00 $\pm$ 0.24 & 38.88 $\pm$ 0.15 & \textbf{69.02 $\pm$ 0.24} & 16.39 $\pm$ 0.13
& 91.43 $\pm$ 0.24 & 52.73 $\pm$ 0.86 & 90.06 $\pm$ 0.04 & 48.82 $\pm$ 0.43
\\
\hline
\end{tabular}%
}
\label{tab:test_baseline}
\end{table*}

\section{Experiments}
\label{sec:exp}
\subsection{Datasets \& Preprocessing}
We evaluate our proposed framework on token-level tasks, including Named Entity Recognition (NER) and Part-of-Speech Tagging (POS) among four languages: Chinese (ZH), Vietnamese (VI), Japanese (JA) and Korean (KO). Based on linguistic structural similarities (SVO vs SOV structural orders) and lexical similarities in terms of phonemic representations, we divide the four languages into 2 groups: (JA,KO) and (ZH,VI) where the former in each pair is considered a high-resourced language and the latter is a low-resourced counterpart. During training, only high-resourced languages are leveraged and we conduct zero-shot evaluation on the target low-resourced languages. 


To evaluate our proposed framework on token-level tasks, we first construct a new dataset by pre-processing the alignment between orthographic and phonemic transcriptions. Specifically, we leverage NER and POS datasets (namely PANX and UDPOS) from XTREME benchmark datasets \cite{hu2020xtreme}. Given input utterances from the datasets, we generate the corresponding phonemic transcriptions on token-level. As phonemic transcriptions can either \textbf{Romanized Transcriptions} (i.e. Pinyin for ZH, Romaji for JA, Romaja for KO) or \textbf{IPA Transcriptions}, we generate both types of phonemic transcriptions and conduct empirical study on both in Section \ref{sec:result}. 

\paragraph{Generating Romanized Transcriptions} As VI is  written in Latin, we preserve the original characters as the corresponding Romanized transcriptions. For ZH, JA, KO, we directly obtain the corresponding Romanized transcriptions via \href{https://dragonmapper.readthedocs.io/en/latest/readme.html}{dragonmapper}, \href{https://pypi.org/project/pykakasi/}{pykakasi} and \href{https://github.com/osori/korean-romanizer}{korean\_romanizer} respectively.  
\paragraph{Generating IPA Transcriptions} As PanPhon \cite{Mortensen-et-al:2016} does not support some of our targeted languages (JA, KO), we leverage external open-source tools to generate IPA transcriptions for individual languages. Specifically, we use \href{https://dragonmapper.readthedocs.io/en/latest/readme.html}{dragonmapper}, \href{https://pypi.org/project/viphoneme/}{viphoneme} to generate IPA transcriptions for ZH, VI respectively. As there are no direct open-source IPA transcription tools available for JA,KO, we generate IPA transcriptions via a two-stage process. First, we generate Romanized transcrptions for JA, KO via \href{https://pypi.org/project/pykakasi/}{pykakasi} and \href{https://github.com/osori/korean-romanizer}{korean\_romanizer} respectively. Then, as the aforementioned Romanized transcriptions are in Latin-based format, we treat them as Latin characters and input them to PanPhon to generate IPA transcriptions for JA and KO. 
Details of our constructed datasets are provided in Table \ref{tab:detailsdataset}.

\subsection{Implementation Details}
\label{subsec:implementation}
For evaluation of both NER and POS tasks, we report the F-1 score for different individual language pairs. We report performance on both development and test set of both source and target languages. 


As IPA involves unique characters that are outside the typical orthographic vocabulary (i.e. \textipa{/N/, /C/, /\textlyoghlig/, /Z/}), we extend PLM vocabulary to account for these special characters. Therefore, both OEs and PEs are resized to account for the newly extended vocabulary. The impact of extended vocabulary will be further discussed in Section \ref{sec:result}.  


For each language pair (ZH-VI vs JA-KO) and token-level tasks (NER vs POS), we tune hyperparameters of our framework based on the development set of the source language. Specifically, we conduct grid search for $\lambda, \beta, \gamma $ over the space [0.1, 0.01, 0.001, 0.0001]. Mask ratio ($\mu$) and cs\_ratio ($r$) are tuned over the space [0.1, 0.4] inclusive with a step of 0.05. Hyperparameter details for each task and language pair are reported  in Table \ref{hyperparameter}.  

We train our model with the batch size of 32 and 16 for mBERT and XLM-R respectively. Both multilingual base versions (L=12, H=12, D=768 where L,H,D denote the number of hidden layers, the number of attention heads per layer and hidden dimension respectively) are used as backbone PLM architectures for our experiments. Both training and inference of our framework are conducted on NVIDIA TITAN RTX and NVIDIA RTX 3090 GPUs. We report our experimental results as the average performance of 3 runs from different random seed initializations with standard deviations. Due to space constraints, we report our empirical studies on test sets in Table \ref{tab:test_full} and \ref{tab:test_baseline}. Additional results on development (dev) sets for both datasets are summarized in the Appendix \ref{sec:app_base}.  

\begin{table}[t]
\centering
\caption{Hyperparameters for PANX and UDPOS datasets (NER and POS tasks respectively) on experimental language pairs ZH->VI and JA->KO}
\resizebox{0.75\columnwidth}{!}{%
\begin{tabular}{|c||c|c||c|c||}
\hline
& \multicolumn{2}{c||}{\textbf{PANX}} & \multicolumn{2}{c||}{\textbf{UDPOS}} \\
\hline 
 & ZH->VI & JA->KO &ZH->VI & JA->KO \\
\hline
 $\lambda$ & 0.01 & 0.1 & 0.01 & 0.1 \\
  $\beta$& 0.01 & 0.001 & 0.001 & 0.01 \\
  $\gamma$ & 0.01 & 0.001 & 0.01 & 0.01 \\
  $\mu$ & 0.20 & 0.25 & 0.10 & 0.05 \\ 
   $r$ & 0.40 & 0.30 & 0.40 & 0.30 \\
\hline
\end{tabular}%
}
\label{hyperparameter}
\end{table}

In our study, we leverage publicly available MUSE bilingual dictionary \cite{conneau2017word}. As bilingual dictionary is only available between EN and target languages, we construct bilingual dictionaries for our language pairs (ZH-VI and JA-KO) by leveraging EN as a bridge for semantic alignment between source and target languages. 
\subsection{Baseline}
We compare our method with previously proposed cross-lingual transfer frameworks for token-level tasks. We conduct experiments with both mBERT and XLM-R backbone PLM architecture. We compare our work with both approaches leveraging Machine Translation (MT) and/or Bilingual Dictionary (Dict), including: 


\begin{itemize}
    \item CoSDA-ML \cite{qin2021cosda}: Multi-level Code-switching augmentation for various cross-lingual NLP tasks, including POS and NER tasks. 
    \item FILTER \cite{fang2021filter}: Cross-lingual Transfer via Intermediate Architecture Disentanglement with Knowledge Distillation objectives from source to target languages. 
    \item xTune \cite{zheng2021consistency}: Two-stage augmentation mechanisms with four exhaustive augmentation methods for cross-lingual transfer. 
    \item XMIXUP \cite{yang2022enhancing}: Cross-lingual transfer via Manifold Mixup Augmentation and Machine Translation Alignment between source and target languages. 
\end{itemize}

As \textit{FILTER, xTune} and \textit{XMIXUP} require training parallel corpora, we generate translation of training data from source languages (ZH, JA) to target languages (VI, KO) via third-party MT package\footnote{https://pypi.org/project/googletrans/}. Ground-truth label sequence of MT data is the same as the original source language data.

As \textit{CoSDA-ML} and \textit{xTune} both require bilingual dictionary for cross-lingual transfer, we leverage the reconstructed MUSE bilingual dictionary for our setting as introduced in \ref{subsec:implementation}. 



    

\section{Result \& Discussion}
\label{sec:result} 


Our experimental results for NER and POS tasks are summarized in Table \ref{tab:test_full} and \ref{tab:test_baseline}. Based on the empirical study, our proposed \textit{PhoneXL} framework consistently outperforms by the backbone PLM architecture of mBERT and XLM-R in both evaluated token-level tasks for low-resourced target languages VI and KO. For instance, with ZH-VI pair, we observe the  target language's F1 evaluation metric improvement of 2.34 points and 2.01 points for NER and POS tasks respectively as compared to the fine-tuned backbone mBERT architecture. This improvement implies the phonemic information provides essential information beyond orthographic representation to further bridge the gap between the source and target languages.

In NER task, the larger and state-of-the-art multilingual PLM XLM-R yields worse performance than mBERT in both of our language pair performance on source and target languages. On the other hand, for POS task, XLM-R based architecture only results in marginal performance gains when compared with mBERT. Interestingly, our mBERT-based framework achieves competitive performance with XLM-R backbone architecture on POS task despite leveraging smaller vocabulary size and less pre-training data. We hypothesize this might be due to the fact that XLM-R has been trained with more languages, leading to biases towards certain types of languages during pre-training that might not share common properties with CJKV languages. 

Despite the performance gain over XLM-R based architecture observed in Table \ref{tab:test_full}, our \textit{PhoneXL} framework
does not consistently outperform the previous baselines such as \textit{FILTER, xTune} and \textit{XMIXUP} in Table \ref{tab:test_baseline}. However, these baselines require extra parallel corpora obtained from machine translation which might not always be readily available for all languages, especially for low-resourced ones. On  the other hand, our proposed method achieves state-of-the-art performance among methods leveraging only bilingual dictionary. In addition, despite its impressive performance, \textit{xTune} requires two-stage training procedures, four different exhaustive augmentation methods as well as the knowledge of both machine translation and bilingual dictionary. Therefore, it is deemed more time-consuming and resource-intensive than our approach. 
\begin{table}[bt]
\centering
\caption{Ablation study of the impact of vocabulary extension and IPA embedding on target language F1-score in POS task  (VI,KO respectively) with mBERT backbone architecture.}
\vspace*{-0.3cm}
\resizebox{\columnwidth}{!}{%
\begin{tabular}{|c||c|c|c|c||}
\hline 
  & \multicolumn{1}{c|}{\textbf{ZH->VI}} & \multicolumn{1}{c|}{\textbf{JA->KO}} \\
 \hline
mBERT (w/o PE, w/o extension) & 48.62 $\pm$ 0.66 & 43.63 $\pm$ 1.28 \\
\hline

mBERT (w PE, w/o extension) &  48.95 $\pm$ 0.52 & 43.85 $\pm$ 0.14 \\
 mBERT (w PE, w extension) & 49.14 $\pm$ 0.98  &  44.42 $\pm$  0.36 \\

PhoneXL-IPA (w/o Lang Embedding) & 49.74 $\pm$ 0.28 & 45.84 $\pm$ 0.07 \\
\hline
PhoneXL-Romanized (full) & 49.78 $\pm$ 1.99  & 45.18 $\pm$ 2.03 \\
\textbf{PhoneXL-IPA (full)} & \textbf{50.71 $\pm$ 0.40}& \textbf{46.75 $\pm$ 0.09} \\
\hline
\end{tabular}%
}
\label{tab:ablation}
\end{table}



\paragraph{Romanized vs IPA Phonemic Transcriptions}
As observed in Table \ref{tab:ablation}, leveraging Romanized transcriptions from individual languages instead of IPA degrades the downstream POS task performance on both low-resourced target languages (averaged 0.93 and 1.57 points of performance drop on VI and KO respectively from \textit{PhoneXL-IPA (full)} to \textit{PhoneXL-Romanized (full)}). We hypothesize it might be due to the lack of phonemic consistency among Romanized transcriptions of different languages. In addition, as certain low-resourced languages might not have their own Romanized transcriptions, IPA phonemic transcriptions provide a more generalized and consistent pathway to generate phonemic transcriptions across different language families.   

\paragraph{Impact of Phonemic Embeddings}
Based on Table \ref{tab:ablation}, we also observe that the introduction of IPA embedding, even without any unsupervised objectives, also provide additional improvements as compared to the backbone orthographic-based mBERT architecture. However, further training with our introduced objectives provide stronger boost in target language performance improvements on the downstream task. 

\paragraph{Impact of Vocabulary Extension}
As observed in Table \ref{tab:ablation}, vocabulary extension is important to learn effective Phonemic Embedding. Vocabulary extension allows the model to differentiate unique IPA characters when encoding tokens, leading to 0.19 and 1.42 points of F1 score improvements on mBERT for VI and KO respectively. It is intuitive since additional special phonemic characters possess different meanings than typical orthographic characters. However, we still observe a significant gap between \textit{mBERT with PE} and \textit{PhoneXL} framework. It is due to the lack of alignment between embedding of phonemic and orthographic inputs.  

\paragraph{Impact of Unsupervised Objectives}
As observed in Table \ref{tab:test_full}, each introduced alignment objective between phonemic and orthographic inputs provides additional performance gain over the original backbone mBERT and XLM-R PLMs on both language groups. Additionally, from Table \ref{tab:ablation}, the existence of performance gap between simple introduction of PE (\textit{mBERT (w PE, w extensions)}) and \textit{PhoneXL} (i.e. 1.57 points on VI and 2.33 points on KO in POS task) implies that the unsupervised alignment objectives are crucial to bringing about the benefits of PE.

\paragraph{Impact of Language Embedding} Language Embedding is crucial to our framework performance, leading to consistent performance gain on both target languages in POS task. In fact, Language Embedding is especially important to $\mathcal{L}_{XMLM}$ as inputs are code-switched sentences which are made up of tokens from different languages. Without language indication from Language Embedding, the model is unable to predict the correct masked tokens in the correct language.

\section{Conclusion \& Future Work}
In our work, we propose \textbf{PhoneXL}, a novel mechanism to integrate phonemic transcription with orthographic transcription to further enhance representation capability of language models in cross-lingual transfer settings. By encouraging the alignment between the two linguistic modalities via direct one-to-one alignment, indirect contextual alignment and additional code-switching via bilingual dictionaries, our proposed \textbf{PhoneXL} yields consistent performance improvements over the backbone orthographic-based PLM architecture in downstream cross-lingual token-level task among the CJKV languages. We also release the first aligned phonemic-orthographic datasets for CJKV languages for two popular token-level tasks (NER and POS). In future work, we plan to train our proposed unsupervised objectives with larger CJKV corpora as pre-training mechanisms to evaluate effectiveness of the representations in multi-granularity downstream tasks (i.e. sentence-level classification tasks to Question-Answering tasks). Further extensions towards few-shot learning settings \cite{nguyen2020dynamic, xia2020cg} where a small number of target language examples can be leveraged to exploit orthographic-phonemic similarity between source and target languages is a promising direction for our future work.

\section*{Limitations}
\label{sec:limit}
Our approach is heavily dependent on the quality of the pre-processed orthographic-phonemic transcription data as it provides the ground-truth for unsupervised alignment objectives. Generating phonemic transcriptions and aligning them correctly with orthographic representations can be costly. Despite our significant efforts, the alignment is still far from perfect optimality. 


Secondly, our approach might not be effective in improving performance on randomly chosen language pairs. As our framework aims to exploit phonemic similarities of languages with different orthographic representations, the methods are only effective in cross-lingual transfer between lexically similar languages in terms of phonology such as CJKV languages. Languages that do not fall into this category might observer little to no performance gains with our proposed framework.

\bibliography{anthology,custom}
\bibliographystyle{acl_natbib}
\balance 
\appendix
\section{Additional Experiments}
\label{sec:app_base}
We provide additional experiments on dev set of PANX and UDPOS datasets of XTREME benchmark datasets in Table \ref{tab:dev_full} and \ref{tab:dev_baseline}. Our observations are mostly consistent between dev and test sets on both evaluated token-level tasks.  

\begin{table*}[htb]
\centering
\caption{NER and POS Experimental Results on PANX and UDPOS \textbf{dev} datasets respectively.}
\vspace*{-0.1cm}
\resizebox{\textwidth}{!}{%
\begin{tabular}{|c||c|c||c|c||c|c||c|c||}
\hline 
 \textbf{Model} & \multicolumn{4}{c||}{\textbf{PANX}} & \multicolumn{4}{c||}{\textbf{UDPOS}}\\
\specialrule{.1em}{0.05em}{.05em}
 & \multicolumn{2}{c||}{ZH->VI} & \multicolumn{2}{c||}{JA->KO} 
  & \multicolumn{2}{c||}{ZH->VI} & \multicolumn{2}{c||}{JA->KO} \\
  \hline 
  & Source (ZH) & \textbf{Target (VI)} & Source (JA) & \textbf{Target (KO)} & Source (ZH) & \textbf{Target (VI)} & Source (JA) & \textbf{Target (KO)} \\
\specialrule{.1em}{0.05em}{.05em}

mBERT 
& 78.54 $\pm$ 0.09 & 49.30 $\pm$ 0.25 & 69.05 $\pm$ 0.12 & 27.27 $\pm$ 0.19
& 92.72 $\pm$ 0.09 & 46.75 $\pm$ 0.66 & 92.42 $\pm$ 0.06 & 42.15 $\pm$ 0.74
 \\
 
\textbf{PhoneXL (full)}
& \textbf{79.89 $\pm$ 0.16} & \textbf{51.81 $\pm$ 0.62} & \textbf{72.20 $\pm$ 0.06} & \textbf{30.00 $\pm$ 0.77} 

& \textbf{93.75 $\pm$ 0.03} & \textbf{49.04 $\pm$ 0.29} & \textbf{96.41 $\pm$ 0.11} & \textbf{44.24 $\pm$ 0.19}  
\\

PhoneXL (w $\mathcal{L}_{align}$) 
& 79.32 $\pm$ 0.24 & 50.78 $\pm$ 0.45 & 71.46 $\pm$ 0.09 & 28.80 $\pm$ 0.40
& 93.52 $\pm$ 0.05 & 48.40 $\pm$ 0.28 & 96.19 $\pm$ 0.06 & 43.65 $\pm$ 0.71
\\

PhoneXL (w $\mathcal{L}_{MLM}$) 
& 79.40 $\pm$ 0.35 & 49.84 $\pm$ 1.72 & 72.12 $\pm$ 0.37 & 27.97 $\pm$ 0.22
& 93.41 $\pm$ 0.13 & 48.70 $\pm$ 0.73 & 96.29 $\pm$ 0.03 & 42.95 $\pm$ 0.58
\\

PhoneXL (w $\mathcal{L}_{XMLM}$) 
& 79.34 $\pm$ 0.04 & 50.55 $\pm$ 0.73 & 72.07 $\pm$ 0.49 & 29.28 $\pm$ 0.80
& 93.43 $\pm$ 0.06 & 48.59 $\pm$ 0.46 & 96.12 $\pm$ 0.08 & 43.18 $\pm$ 0.70
\\
\hline
\hline
XLM-R
& 75.33 $\pm$ 0.71 & 35.77 $\pm$ 0.75 & 66.14 $\pm$ 0.016 & 14.56 $\pm$ 0.80
& 94.92 $\pm$ 0.08 & 48.96 $\pm$ 0.46 & 96.44 $\pm$ 0.08 & 44.15 $\pm$ 0.17
 \\
\textbf{PhoneXL (full)}
& \textbf{76.54 $\pm$ 0.16} & \textbf{38.90 $\pm$ 0.31} & \textbf{68.85 $\pm$ 0.16} & \textbf{17.26 $\pm$ 0.23}
& 94.84 $\pm$ 0.41 & 51.45 $\pm$ 1.55 & \textbf{97.18 $\pm$ 0.02} & \textbf{45.97 $\pm$ 0.48}
\\

PhoneXL (w $\mathcal{L}_{align}$) 
& 76.27 $\pm$ 0.24 & 36.42 $\pm$ 0.23 & 68.17 $\pm$ 0.14 & 16.08 $\pm$ 0.01 
& \textbf{95.00 $\pm$ 0.05} & \textbf{51.47 $\pm$ 0.30} & 96.67 $\pm$ 0.08 & 45.00 $\pm$ 0.44
\\

PhoneXL (w $\mathcal{L}_{MLM}$) 
& 76.16 $\pm$ 0.10 & 37.20 $\pm$ 0.11 & 68.12 $\pm$ 0.15 & 15.90 $\pm$ 0.35 
& 94.26 $\pm$ 0.17 & 50.84 $\pm$ 1.56 & 96.70 $\pm$ 0.04 & 45.27 $\pm$ 0.41
\\
PhoneXL (w $\mathcal{L}_{XMLM}$) 
& 76.09 $\pm$ 0.06 & 36.81 $\pm$ 0.28 & 68.26 $\pm$ 1.27 & 16.11 $\pm$ 0.23 
& 94.22 $\pm$ 0.14 & 50.48 $\pm$ 1.06 & 96.67 $\pm$ 0.05 & 44.88 $\pm$ 0.62
\\
\hline
\end{tabular}%
}
\label{tab:dev_full}
\end{table*}

\begin{table*}[htb]
\centering
\caption{NER and POS Baseline Results on PANX and UDPOS \textbf{dev} datasets respectively. \textbf{Dict} denotes the assumptions of available bilingual dictionary and \textbf{MT} refers to the assumptions of available Machine Translations between source and target languages. Cross-lingual Transfer methods leverage either Dict or MT or both.  }
\vspace*{-0.1cm}
\resizebox{\textwidth}{!}{%
\begin{tabular}{|c||c|c||c|c||c|c||c|c||c|c||}
\hline 
  \textbf{Model} & \multicolumn{2}{c||}{\textbf{Assumption}} & \multicolumn{4}{c||}{\textbf{PANX}} & \multicolumn{4}{c||}{\textbf{UDPOS}}\\
\specialrule{.1em}{0.05em}{.05em}
 &  Dict & MT & \multicolumn{2}{c||}{ZH->VI} & \multicolumn{2}{c||}{JA->KO} 
  & \multicolumn{2}{c||}{ZH->VI} & \multicolumn{2}{c||}{JA->KO} \\
  \hline 
  & & &  Source (ZH) & \textbf{Target (VI)} & Source (JA) & \textbf{Target (KO)} & Source (ZH) & \textbf{Target (VI)} & Source (JA) & \textbf{Target (KO)} \\
\specialrule{.1em}{0.05em}{.05em}

mBERT 
& & 
& 78.54 $\pm$ 0.09 & 49.30 $\pm$ 0.25 & 69.05 $\pm$ 0.12 & 27.27 $\pm$ 0.19
& 92.72 $\pm$ 0.09 & 46.75 $\pm$ 0.66 & 92.42 $\pm$ 0.06 & 42.15 $\pm$ 0.74
 \\

CoSDA-ML 
& $\checkmark$ & 
& 78.06 $\pm$ 0.19 & 47.22 $\pm$ 1.39 & 70.46 $\pm$ 0.43 & 26.10 $\pm$ 1.57
& 92.97 $\pm$ 0.17 & 48.46 $\pm$ 0.59 & 94.66 $\pm$ 0.18 & 40.85 $\pm$ 0.91
\\

X-MIXUP 
& & $\checkmark$
& 78.68 $\pm$ 0.21 & \textbf{53.62 $\pm$ 0.45} & 67.70 $\pm$ 0.58 & 26.70 $\pm$ 0.87 
&  \textbf{94.26 $\pm$ 0.23} & 48.61 $\pm$ 0.62 & 96.02 $\pm$ 0.07 &  \textbf{48.70 $\pm$ 0.51} 
\\
\hline

\textbf{PhoneXL (full)} & \checkmark & 
& \textbf{79.89 $\pm$ 0.16} & 51.81 $\pm$ 0.62 & \textbf{72.20 $\pm$ 0.06} & \textbf{30.00 $\pm$ 0.77} 

& 93.75 $\pm$ 0.03 & \textbf{49.04 $\pm$ 0.29} & \textbf{96.41 $\pm$ 0.11} & 44.24 $\pm$ 0.19 
\\
\hline
\hline
XLM-R
& &
& 75.33 $\pm$ 0.71 & 35.77 $\pm$ 0.75 & 66.14 $\pm$ 0.016 & 14.56 $\pm$ 0.80
& 94.92 $\pm$ 0.08 & 48.96 $\pm$ 0.46 & 96.44 $\pm$ 0.08 & 44.15 $\pm$ 0.17
\\
FILTER 
& & $\checkmark$
& 72.13 $\pm$ 0.16 & 39.76 $\pm$ 1.31 & 62.96 $\pm$ 0.38 & 19.46 $\pm$ 0.95 
& 93.14 $\pm$ 0.19 & \textbf{53.89 $\pm$ 0.28} & 96.99 $\pm$ 0.07 & 39.39 $\pm$ 1.56
\\

xTune 
& $\checkmark$ & $\checkmark$
& \textbf{77.38 $\pm$ 0.10} & \textbf{40.58 $\pm$ 0.67} & 68.26 $\pm$ 0.38 & \textbf{23.05 $\pm$ 0.95} 

& \textbf{95.34 $\pm$ 0.17} & 50.29 $\pm$ 0.69 & 97.08 $\pm$ 0.08 & \textbf{49.54 $\pm$ 0.79}
\\

X-MIXUP 
& & $\checkmark$
&  73.20 $\pm$ 0.22 & 38.17 $\pm$ 0.78  & 64.46 $\pm$ 0.47 & 17.00 $\pm$ 2.81

&  94.80 $\pm$ 0.14 & 50.25 $\pm$ 1.05   &  96.42 $\pm$ 0.05 &  48.05 $\pm$ 0.96
\\

\hline 
\textbf{PhoneXL (full)}
& $\checkmark$ &
& 76.54 $\pm$ 0.16 & 38.90 $\pm$ 0.31 & \textbf{68.85 $\pm$ 0.16} & 17.26 $\pm$ 0.23
& 94.84 $\pm$ 0.41 & 51.45 $\pm$ 1.55 & \textbf{97.18 $\pm$ 0.02} & 45.97 $\pm$ 0.48
\\
\hline
\end{tabular}%
}
\label{tab:dev_baseline}
\end{table*}



\end{document}